\newif\ifRAL
\RALfalse 	
\newif\ifTR
\TRfalse 	
\newif\ifPrePrint
\PrePrintfalse

\newif\ifDraft
\Draftfalse

\ifRAL
\documentclass[letterpaper, 10 pt, journal, twoside]{IEEEtran}
\else
\documentclass[letterpaper, 10 pt, conference]{ieeeconf}
\fi

\IEEEoverridecommandlockouts

\ifRAL
\else
\fi		

\makeatletter

\let\proof\@undefined
\let\endproof\@undefined
\makeatother

\usepackage{amsmath} \usepackage{amssymb}  \usepackage{amsthm}
\usepackage{amsfonts}
\usepackage{mathtools}
\usepackage{acro}
\usepackage{bm}
\usepackage{cite}
\usepackage{layouts}
\providecommand{\bm}{\pmb}

\theoremstyle{definition}

\theoremstyle{remark}

\usepackage{verbatim}

\usepackage{color}
\usepackage{subcaption}
\usepackage{graphicx}
\usepackage{caption}
\usepackage{siunitx}

\usepackage{optidef}

\usepackage{times}

\usepackage{xspace}

\usepackage[ruled,vlined,linesnumbered]{algorithm2e}
\usepackage[noend]{algorithmic}

\graphicspath{{images/}}

\usepackage[us]{datetime}
\usepackage{caption}
\usepackage[usenames,dvipsnames,table]{xcolor}

\usepackage{hyperref} \hypersetup{
    colorlinks,
    citecolor=black,
    filecolor=black,
    linkcolor=black,
    urlcolor=black,
    pdfauthor={},
    pdfsubject={},
    pdftitle={}
}
\usepackage[capitalise]{cleveref}
\crefformat{equation}{(#2#1#3)}
\let\labelindent\relax
\usepackage[inline]{enumitem}

\usepackage{graphicx}

\usepackage{latexsym}
\usepackage{color}

\usepackage{cite}

\usepackage{caption}

\usepackage{tikz}
\usetikzlibrary{shapes,arrows}
\usetikzlibrary{arrows.meta}
\usetikzlibrary{positioning,fit,backgrounds}

\usepackage{tabularx, booktabs}
\newcolumntype{Y}{>{\centering\arraybackslash}X}
\usepackage{multirow}

\usepackage{acro}
\usepackage{bm}
\usepackage{mathtools}
\usepackage{amsfonts}
\usepackage{amsmath}
\usepackage{subcaption}
\usepackage{xcolor}
\usepackage{siunitx}
\usepackage{hyperref}
\usepackage{svg}
\usepackage{optidef}

\DeclareAcronym{ASL}{short = ASL, long = Autonomous Systems Lab}
\DeclareAcronym{OMAV}{short = OMAV, long = Omnidirectional Micro Aerial Vehicle}
\DeclareAcronym{MAV}{short = MAV, long = Micro Aerial Vehicle}
\DeclareAcronym{DOF}{short = DOF, long = degrees of freedom}
\DeclareAcronym{PBC}{short = PBC, long = passivity-based control}
\DeclareAcronym{PH}{short = PH, long = Port-Hamiltonian}
\DeclareAcronym{NDT}{short = NDT, long = non-destructive testing}
\DeclareAcronym{PEMS}{short = PEMS, long = Power and Energy Monitoring System}
\DeclareAcronym{WTC}{short = WTC, long = wrench tracking controller}
\DeclareAcronym{PTC}{short = PTC, long = pose tracking controller}
\DeclareAcronym{MBE}{short = MBE, long = momentum-based wrench estimator}
\DeclareAcronym{ASIC}{short = ASIC, long = Axis-Selective Impedance Control}
\DeclareAcronym{MPC}{short = MPC, long = Model Predictive Control}
\DeclareAcronym{MPPI}{short = MPPI, long = Model Predictive Path Integral}
\DeclareAcronym{APhI}{short = APhI, long = Aerial Physical Interaction}
\DeclareAcronym{LLE}{short = LLE, long = Largest Lyapunov Exponent}
\DeclareAcronym{ICBF}{short = ICBF, long = Integral Control Barrier Function}
\DeclareAcronym{CBF}{short = CBF, long = Control Barrier Function}
\DeclareAcronym{COM}{short = CoM, long = Center of Mass}
\DeclareAcronym{AM}{short = AM, long = Aerial Manipulator}
\DeclareAcronym{HRI}{short = HRI, long = Human Robot Interaction}
\DeclareAcronym{RL}{short = RL, long = Reinforcement Learning}
\DeclareAcronym{PPO}{short = PPO, long = Proximal Policy
	Optimization}
\DeclareAcronym{DoF}{short = DoF, long = degree of freedom, short-plural = s, long-plural-form = degrees of freedom}
\DeclareAcronym{FT}{short = F/T, long = force and torque, short-indefinite = an, long-indefinite = a}

\newcommand{\obsVec}{\bm{o}}
\newcommand{\actVec}{\bm{a}}
\newcommand{\reward}{r}

\newcommand{\discount}{\gamma}
\newcommand{\doorAngle}{\alpha}

\newcommand{\rhandle}{r_{\text{h}}}
\newcommand{\rhandleDist}{r_{\text{h\_dist}}}
\newcommand{\rhandleIn}{r_{\text{h\_in}}}
\newcommand{\dhandleTresh}{\delta_{\text{h\_tresh}}}

\newcommand{\rattitude}{r_{\text{att}}}
\newcommand{\attErr}{\theta_{hd}}

\newcommand{\rdoor}{r_{\text{d}}}
\newcommand{\rdoorDist}{r_{\text{d\_dist}}}
\newcommand{\rdoorOpen}{r_{\text{d\_open}}}
\newcommand{\ddoorTresh}{\delta_{\text{d\_tresh}}}

\newcommand{\rvel}{r_{\text{vel}}}
\newcommand{\rvelLin}{r_{\text{vel,lin}}}
\newcommand{\rvelAng}{r_{\text{vel,ang}}}

\newcommand{\rwrench}{r_{\tau}}

\renewcommand{\vec}[1]{\bm{#1}}		
\renewcommand{\(}{\left(}		
\renewcommand{\)}{\right)}		
\renewcommand{\[}{\left[}		
\renewcommand{\]}{\right]}		
\newcommand{\matr}[1]{\bm{#1}}		
\newcommand{\nR}[1]{\mathbb{R}^{#1}}		
\newcommand{\SO}[1]{\mathsf{SO}(#1)}		
\newcommand{\modulus}[1]{\left| #1 \right|}	
\newcommand{\upperRomannumeral}[1]{\uppercase\expandafter{\romannumeral#1}}	

\newcommand{\vSpace}{\;\,}
\newcommand{\transpose}{^\top}
\newcommand{\refer}{^{\text{ref}}}
\newcommand{\W}[1]{\prescript{}{W}{#1}}
\newcommand{\B}[1]{\prescript{}{B}{#1}}
\newcommand{\D}[1]{\prescript{}{D}{#1}}

\renewcommand{\frame}[1]{\mathcal{F}_{#1}}		

\newcommand{\pos}{\vec{p}}				
\newcommand{\vel}{\vec{v}}				
\newcommand{\acc}{\dot{\vel}}				

\newcommand{\rotMat}{\matr{R}}				

\newcommand{\frameW}{\frame{W}}			
\newcommand{\frameB}{\frame{B}}			
\newcommand{\frameD}{\frame{D}}			
\newcommand{\frameH}{\frame{H}}			

\newcommand{\rotMatWB}{\rotMat_{WB}}	
\newcommand{\rotMatWD}{\rotMat_{WD}}	
\newcommand{\angVel}{\vec{\omega}}

\DeclarePairedDelimiter{\norm}{\lVert}{\rVert} 

\newcommand{\wrench}{\bm{\tau}}

\newcommand{\totalInertia}{\bm{M}}

\newcommand{\Coriolis}{\bm{c}}

\newcommand{\wrenchGravity}{\bm{g}}

\newcommand{\wrenchCommand}{\wrench_{c}}

\newcommand{\wrenchPosContr}{\wrench_{p}}

\newcommand{\pdot}{\dot{\bm{p}}}

\newcommand{\Uniform}[2]{\mathcal{U}\left(#1, #2\right)}

\ifRAL \fi

\author{Eugenio Cuniato$^1$, Ismail Geles$^1$, Weixuan Zhang$^1$, Olov Andersson$^1$, Marco Tognon$^2$, Roland Siegwart$^1$
	\thanks{${}^1$ Autonomous Systems Lab (ASL), ETH Zurich.}
        \thanks{${}^2$ Univ Rennes, CNRS, Inria, IRISA, Campus de Beaulieu, 35042 Rennes Cedex, France.} 
        \thanks{Corresponding author: {\tt \footnotesize \href{mailto:ecuniato@ethz.ch}{ecuniato@ethz.ch}.}}
	\thanks{The research leading to these results has been supported by the AERO-TRAIN project, European Union's Horizon 2020 research and innovation program under the Marie Skłodowska-Curie grant agreement No 953454. The authors are solely responsible for its content.}
}

\ifRAL \title{Learning to Open Doors with an Aerial Manipulator}

\else \title{\bf Learning to Open Doors with an Aerial Manipulator}
\fi

\ifPrePrint
\makeatletter
\def\ps@titlepagestyle{
	\def\@oddfoot{}\def\@evenfoot{}
	\def\@oddhead{\textcolor{red}{\sf\footnotesize Preprint version, final version at http://ieeexplore.ieee.org/ \hfill IEEE Robotics and Automation Letters 2022}}
	\def\@evenhead{\textcolor{red}{\sf\footnotesize  Preprint version, final version at http://ieeexplore.ieee.org/  \hfill IEEE Robotics and Automation Letters 2022}}}\makeatother
\makeatletter
\def\ps@headings{
	\def\@oddfoot{\textcolor{red}{\sf\footnotesize  Preprint version, final version at http://ieeexplore.ieee.org/ \hfill \thepage \;\;~\hfill~\hfill IEEE Robotics and Automation Letters 2022}}\def\@evenfoot{\hfill\thepage\hfill}
	\def\@oddhead{}\def\@evenhead{}}\makeatother
\fi	

\ifDraft
\makeatletter
\def\ps@titlepagestyle{
	\def\@oddfoot{}\def\@evenfoot{}
	\def\@oddhead{\textcolor{red}{\sf Draft version  \hfill Confidential}}
	\def\@evenhead{\textcolor{red}{\sf  Draft version  \hfill Confidential}}}\makeatother
\makeatletter
\def\ps@headings{
	\def\@oddfoot{\textcolor{red}{\sf  Draft version  \hfill Confidential}}\def\@evenfoot{\hfill\thepage\hfill}
	\def\@oddhead{}\def\@evenhead{}}\makeatother
\usepackage{draftwatermark}
\SetWatermarkText{Confidential draft}
\SetWatermarkScale{0.6}
\SetWatermarkLightness{0.9} 

\fi

\begin{document}

\maketitle

\begin{abstract}
	The field of aerial manipulation has seen rapid advances, transitioning from push-and-slide tasks to interaction with articulated objects. 	The motion trajectory of these complex actions is usually hand-crafted or a result of online optimization methods like \ac{MPC} or \ac{MPPI} control. 	However, these methods rely on heuristics or model simplifications to efficiently run on onboard hardware, limiting their robustness, and making them sensitive to disturbances and differences between the real environment and its model. 	In this work, we propose a \ac{RL} approach to learn reactive motion behaviors for a manipulation task while producing policies that are robust to disturbances and modeling errors. 	Specifically, we train a policy to perform a door-opening task with an \ac{OMAV}. 	The policy is trained in a physics simulator and shown in the real world, where it is able to generalize also to door closing tasks never seen in training.
	We also compare our method against a state-of-the-art \ac{MPPI} solution in simulation, showing a considerable increase in robustness and speed.
\end{abstract}

\ifRAL 		
	\else 	{} \fi

\section{Introduction}
Aerial robotics is a growing area of research developing innovative systems for a wide range of applications~\cite{Ollero, RuggieroReview}. The field was particularly boosted by the development of novel \acp{MAV}~\cite{2021f-HamUsaSabStaTogFra} that are fully actuated~\cite{8485627, Tognon2019, Ryll20196DIC} or omnidirectional~\cite{9295362}, capable of generating a six \acp{DoF} control wrench independently of their attitude.
This decoupling enabled many new aerial physical interaction tasks otherwise very difficult for standard multirotor systems and reduced the need for complex end-effectors to compensate for the motion of the body.

Such aerial physical interaction tasks are especially relevant in domains such as industrial inspection and maintenance, where many tasks require complex interaction between the robot and the structure of interest.
While most current investigations are limited to pure touching or push-and-slide operations with static surfaces~\cite{Rashad2019,nava2019direct}, recent works have tackled the problem of interaction with non-static objects, like pushing a cart with possibly unknown properties~\cite{9812342,cuniato2022power,benzi2022adaptive}. When more complex actions are required, the motion trajectory is usually hand-crafted and dependent on heuristics. Examples of this include the door opening task performed in~\cite{tsukagoshi2015aerial} or the valve turning in~\cite{korpela2014towards}. One way of generating optimized motion trajectories was presented in~\cite{lee2020aerial}, where \ac{MPC} was employed to open a door with an aerial manipulator. However, these methods require an analytical description of the environment, the robot, and most importantly the contact dynamics. The latter are often nonlinear and can easily increase the optimization problem's complexity if accurate simplifications are not found. \begin{figure}[t]
	\centering
	\includegraphics[width=\columnwidth]{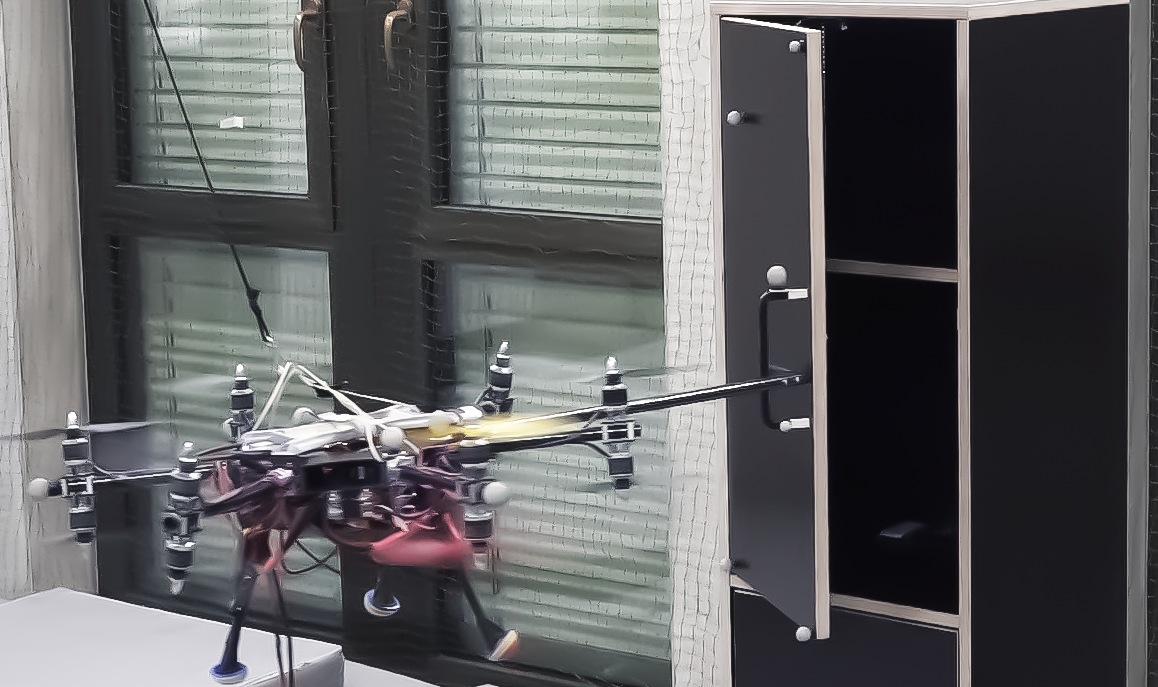}
	\caption{The aerial robot opening a cabinet door during real-world experiments. The robot learns to both open and close the door in real life, even though it was only trained on opening in simulation.}
	\label{fig:omav}
\end{figure}

To overcome these complexities, sampling-based approaches like \ac{MPPI}~\cite{williams2017information} have been receiving attention since they solve optimization problems by recreating the desired task in a dynamic simulator rather than by constructing a full analytical model. This technique was successfully employed in robotic manipulation tasks~\cite{abraham2020model}, drone racing~\cite{lee2020aggressive}, autonomous driving~\cite{williams2017information}, and more recently in an autonomous aerial door-opening and valve-turning tasks~\cite{brunner2022planning}. However, while this approach can handle complex contact dynamics while reducing the engineering effort, the sampling can be computationally very expensive since it requires running multiple dynamic simulations online in parallel. This reduces the number of samples, practically reducing the chances of generating feasible and goal-accomplishing trajectories. Moreover, this approach is very sensitive to uncertainties and differences between the simulated and the real world, which limits its applicability in uncertain environments.

In more traditional, non-aerial manipulation literature, \acl{RL} has received considerable attention for a variety of complex tasks.Among the various examples, in~\cite{gu2017deep} a robotic arm learned different policies to reach a target, relocate objects, and open doors, and in~\cite{rajeswaran2017learning} a 27 \ac{DoF} robotic hand learned complex manipulation skills like hammering a nail or turning and pulling a door's handle. While \ac{RL} has many similarities with \ac{MPPI},
it provides two main advantages: \textit{(i)} the learning process can be performed offline and offboard on more powerful hardware, generating a policy that is fast to infer online and\textit{(ii)} the uncertainties in the robot's or environment's model, as well as external disturbances, can be embedded into the training process through \textit{domain randomization}, leading to robust policies. This was also demonstrated in the aerial community for high-speed and acrobatic flying with quadrotors~\cite{hwangbo2017control, loquercio2019deep, song2021autonomous}. However, despite its potential, this approach is so far rarely adopted in the aerial manipulation field. A few examples include aerial transportation of suspended loads~\cite{faust2017automated} and, very recently, push-and-slide tasks \cite{zhang2022learning}, but aerial manipulation of articulated objects has not yet been explored using \ac{RL}.

\subsection{Methodology and Contributions}
In this work, we explore the novel use of \ac{RL} for aerial physical interaction with articulated objects, focusing on a door opening task with an \ac{OMAV} (depicted in~\cref{fig:omav}). This is a particularly challenging example for two reasons: the \ac{OMAV} features complex and slow tilt-arm dynamics which are difficult to simulate (complicating sim-to-real transfer), and the task itself requires centimeter-precise execution in the presence of aerodynamic disturbances from flying close to objects.

We believe that this makes the door opening task a great example to investigate the strengths and weaknesses of \ac{RL}-based approaches for complex aerial manipulation tasks.

We train a reactive motion control policy in simulation, which provides pose references to a lower-level pose controller, transfer it to the real world, and compare it to a state-of-the-art model-based control method for the same task.

Specifically, we offer the following contributions:
\begin{itemize}
	\item We explore the use of \ac{RL} for a complex aerial manipulation task, specifically door-opening.
	\item We show that the learned policy outperforms a state-of-the-art \ac{MPPI} controller in terms of robustness to errors environment observations and changes in initial conditions as well as task completion speed.
	\item We perform sim-to-real transfer on the policy and analyze its behavior on the real robot, including desirable emergent properties like ability to recover from initial failed attempts and generalization to new variants of the task, such as door closing.
\end{itemize}

\section{Model and Control}
Our goal is to learn a reactive motion policy for door opening with an \acl{AM} composed of an \acl{OMAV} and a rigid stick with a hook.
The policy outputs a desired body pose to the pose controller of the robot. In this section we describe the physical model of the \ac{OMAV} and the door.
\subsection{Aerial robot}

\subsubsection*{Modeling}
\label{sec:AMmodeling}
The aerial robot is an \ac{OMAV} composed of a rigid body with six independently tiltable rotor arms and twelve propellers.
Controlling both arm angles and propeller speed, the platform is fully actuated in any orientation, i.e., it can always exert a full 6-\ac{DoF} wrench.

We describe the position and orientation of the robot's body frame $\frameB$ with respect to the world frame $\frameW$ by the vector $\W{\pos} \in \nR{3}$ and the rotation matrix $\rotMatWB  \in \SO{3}$, respectively.
Then, as already in~\cite{9295362,Brunner2020}, the robot's Lagrangian model can be expressed as
\begin{equation}
	\label{omav_original_model}
	\totalInertia \B\acc + \Coriolis(\B\angVel) + \wrenchGravity(\rotMatWB) = \wrenchPosContr,
\end{equation}
with inertia matrix $\totalInertia \in \nR{6\times 6}$, control wrench $\wrenchPosContr \in \nR{6}$, gravity wrench $\wrenchGravity(\rotMatWB) \in \nR{6}$, centrifugal and Coriolis wrench $\Coriolis(\B\angVel) \in \nR{6}$ and $\B\vel=\left[ \B\pdot\transpose \vSpace \B\angVel\transpose\right]\transpose \in \nR{6}$ system's twist all expressed in the body frame.
The control wrench $\wrenchCommand$ is mapped into the \ac{OMAV}'s arm tilt angles and rotor speeds as in \cite{9295362}. To enable the system to interact with the environment, it is equipped with a rigid stick ending with an hook.
We consider a hook frame $\frameH$ attached in the geometric middle of the hook cylinder (\cref{fig:omav-frames}) with the same orientation of the \ac{OMAV}'s body $\rotMatWB$.

\subsubsection*{Pose controller}
\label{sec:pose_control}
We use a pose controller to steer the vehicle's pose toward a reference position $\W{\pos}\refer \in \nR{3}$ and orientation $\rotMatWB\refer{} \in \SO{3}$, using the control wrench $\wrenchCommand$. For more details about the controller, we refer the reader to~\cite{9295362}.

\subsection{Door model}
The door is composed of two rigid bodies attached through a 1-\ac{DoF} rotational joint at the door's hinge. The door's handle is rigidly attached on its right side. The robot can fit its hook inside the handle to swing the door open, as shown in \Cref{fig:omav}. We define the pose of the door's handle through the frame $\frameD$ located in the center of the handle's rectangle (\cref{fig:omav-frames}). We also denote its position and orientation in $\frameW$ by the vector $\W{\pos}_d \in \nR{3}$ and the rotation matrix $\rotMatWD  \in \SO{3}$, respectively. Finally, we denote the angle of the door w.r.t. its fixed base with $\doorAngle \in \left[0,\frac{\pi}{2}\right]\SI{}{\radian}$, where $\doorAngle=\frac{\pi}{2}$ indicates open and  $\doorAngle=0$ indicates closed.

\section{Policy Description and Training}
\begin{figure}[t]
	\centering
	\centerline{\includegraphics[width=0.9\columnwidth]{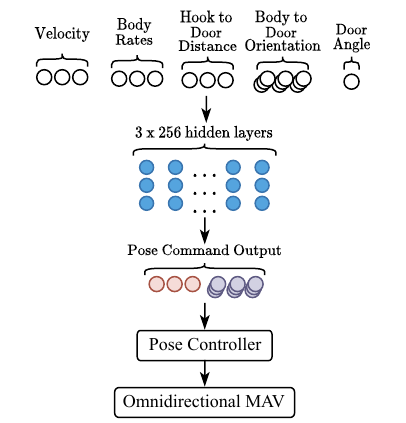}}
	\caption{The robot control scheme composed of the learned door opening policy and the model-based pose controller. The policy knows the door's opening angle, the relative pose w.r.t. the door and the current robot's velocity. It then outputs a desired pose for the robot to follow in order to open the door.}
	\label{fig:control_scheme}
\end{figure}
Consider an \ac{RL} scenario in which a policy receives an observation vector $\obsVec_k \in \nR{n_o}$ at each time step $k$, producing an action vector $\actVec_k \in \nR{n_a}$. The action affects the simulated environment and gets evaluated with a reward function $\reward_k \left(\obsVec_k,\actVec_k\right) \in \nR{}$. The agent's goal is to maximize the discounted sum of rewards $\sum_{k=0}^{\infty}\discount^k \reward_k$, with discount factor $\discount < 1$ affecting the effective time horizon over which the agent plans. An overview of our policy's observations and actions is in \cref{fig:control_scheme}.

\subsection{Observation space}
The policy's observations are given by a 19-dimensional vector containing information about both the robot and door state, composed of the following:

\subsubsection{Robot velocity}
The linear $\B\vel_p \in \nR{3}$ and angular $\B\angVel \in \nR{3}$ velocities of the robot's body in body frame.
\subsubsection{Hook-to-door distance}
The distance between the hook frame $\frameH$ and the door handle frame $\frameD$, where the hook should fit to pull the door, defined as $\B\pos_{hd}= \B\pos_h - \B\pos_d$.
\subsubsection{Body-to-door orientation}
The flattened rotation matrix $\rotMat_{BD}$ representing the orientation between the body frame $\frameB$ and the door frame $\frameD$.\subsubsection{Door angle}
The door's hinge angle $\doorAngle \in \left[0,\frac{\pi}{2}\right]$.\subsection{Action space}
The policy outputs the desired pose corrections into the door frame $\frameD$ for the robot, encoded in the action vector $\actVec = \left[\D\pos\refer \; \vec{\lambda}_0 \; \vec{\lambda}_1\right]$. The desired position command for the inner pose controller is then obtained as $\W\pos\refer = \W\pos + \rotMatWD\D\pos\refer$. The two vectors $\vec{\lambda}_0$,$\vec{\lambda}_1 \in \nR{3}$ represent two axes of the desired commanded rotation frame.
This allows us to use a continuous representation of the orientation space while using only two 3-dimensional vectors instead of a full 9-dimensional rotation matrix. We then compute the correction rotation matrix $\rotMat\refer = \[\Vec{e}_0 \;\; \Vec{e}_1 \;\; \Vec{e}_0 \times \Vec{e}_1\]$, applying the Gram-Schmidt orthogonalization method to the vectors $\vec{\lambda}_0$,$\vec{\lambda}_1$ as
		\begin{equation}
			\label{gram_schmidt}
			\Vec{e}_0  = \frac{\vec{\lambda}_0}{\norm{\vec{\lambda}_0}}, \;\;
			\Vec{e}_1 = \frac{\vec{\lambda}_1 - \(\Vec{e}_0 \cdot \vec{\lambda}_1\)\Vec{e}_0}{\norm{\vec{\lambda}_1 - \(\Vec{e}_0 \cdot \vec{\lambda}_1\)\Vec{e}_0}},
		\end{equation}
		where $\Vec{e}_0 \cdot \vec{\lambda}_1 \in \nR{}$ is the dot product between the two vectors. 				The desired attitude reference is then computed as $\rotMatWB\refer = \rotMatWD\rotMat\refer$. 		This way, the policy is commanding pose displacements w.r.t. the current pose - and, implicitly, velocities - to the \ac{AM}'s pose controller. 		The bigger the produced offset, the faster the motion of the \ac{AM} in the desired direction.

		\subsection{Reward function}
		\begin{figure}[t]
			\centering
			\includegraphics[width=\columnwidth]{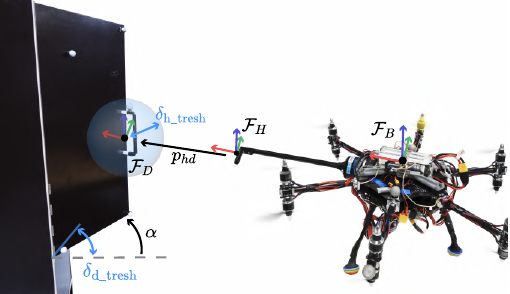}
			\caption{Visual definition of frames and reward quantities. The goal of the policy is to first line up $\frameH$ and $\frameD$ and then open the door by driving the door angle, $\alpha$, to zero.}
			\label{fig:omav-frames}
		\end{figure}
		\begin{table}[b]
			\renewcommand*{\arraystretch}{1.4}
			\centering
			\begin{tabular}{|| c | c  c | c  c||}
				\hline
				Reward                      & \multicolumn{2}{c}{Reward definition} \vline & \multicolumn{2}{c}{Scaling coefficient}  \vline\;\vline                  \\
				\hline\hline
				\multirow{2}{*}{$\rhandle$} & $\rhandleDist$                               & $-\norm{\B\pos_{hd}}$                                   & $c_0$ & $1000$ \\
				                            & $\rhandleIn$                                 & $\rhandleDist > \dhandleTresh$                          & $c_1$ & $1000$ \\
				\hline
				$\rattitude$                & $\rattitude$                                 & $-\attErr$                                              & $c_2$ & $1000$ \\
				\hline
				\multirow{2}{*}{$\rdoor$}   & $\rdoorDist$                                 & $-\modulus{\doorAngle}$                                 & $c_3$ & $100$  \\
				                            & $\rdoorOpen$                                 & $\doorAngle > \ddoorTresh$                              & $c_4$ & $2000$ \\
				\hline
				\multirow{2}{*}{$\rvel$}    & $\rvelLin$                                   & $-\norm{\B\pdot}^2$                                     & $c_5$ & $10$   \\
				                            & $\rvelAng$                                   & $-\norm{\B\angVel}^2$                                   & $c_6$ & $10$   \\
				\hline
				$\rwrench$                  & $\rwrench$                                   & $-\norm{\wrenchPosContr}^2$                             & $c_7$ & $1$    \\
				\hline
			\end{tabular}
			\caption{Definition of the five main reward components and their coefficients to guide the policy towards the door opening motion.\label{tab:rewards}}
		\end{table}
		The ultimate goal of the policy is to open the door (rotate it clockwise by $90^\circ$). 		To correctly and safely perform this task, we encode two requirements into the reward function: \textit{(i)} we want to open the door by pulling or pushing the door's handle with the robot's hook; \textit{(ii)} we would like to keep the hook frame ($\frameH$) as perpendicular as possible to the door, which maintains the maximum safety distance between the robot's body and the door. 		We encode these requirements into the reward function
		\begin{equation}
			\label{reward_function}
			\reward = \rhandle + \rattitude + \rdoor + \rvel + \rwrench ,
		\end{equation}
		whose elements are described below and graphically represented in~\cref{fig:omav-frames}. \cref{tab:rewards} contains the reward definitions and weights used for training the policy.
		\subsubsection{Hook in door's handle ($\rhandle$)}
		This reward pushes the hook towards the door handle, defined as $\rhandle = c_0 \rhandleDist + c_1 \rhandleIn$. Here $\rhandleDist = -\norm{\B\pos_{hd}}$ gives a reward based on how close the hook is to the door's handle, while
		\begin{equation*}
			\rhandleIn =
			\begin{cases}
				-1 & \text{if}\; \rhandleDist > \dhandleTresh \\
				0  & \text{if}\; \rhandleDist < \dhandleTresh
			\end{cases}
			,
		\end{equation*}
		gives an additional reward if the hook is inside the handle, creating a region of attraction with radius $\dhandleTresh=\SI{0.06}{\meter}$.
		\subsubsection{Body-to-door angle ($\rattitude$)}
		To avoid unnecessary torsion of the robot's body during the hooking or opening phase, we wish to keep the hook perpendicular to the plane of the door (i.e., align the attitudes of the frames $\frameD$ and $\frameH$). 		We define an attitude reward $\rattitude = -c_2 \attErr$, where $\attErr = \cos^{-1}{0.5\left(Tr\left(\rotMat_{BD}\right)-1\right)}$ represents the angular distance of the rotation $\rotMat_{BD}$, with $Tr\left(\rotMat_{BD}\right)$ the trace of the matrix. 		This way, $\attErr=0$ if the attitudes of the hook $\frameH$ and the door $\frameD$ frames are aligned, or $\attErr>0$ otherwise. 		\subsubsection{Door opening ($\rdoor$)}
		This reward makes the robot pull the door open ($\doorAngle=0$). 		It is given by $\rdoor = c_3 \rdoorDist + c_4 \rdoorOpen$, where $\rdoorDist=-\modulus{\doorAngle}$ pushes the door towards the target and
		\begin{equation*}
			\rdoorOpen =
			\begin{cases}
				-1 & \text{if}\; \doorAngle > \ddoorTresh \\
				0  & \text{if}\; \doorAngle < \ddoorTresh
			\end{cases}
			,
		\end{equation*}
		gives an additional reward if the robot opens the door of at least $\doorAngle<\ddoorTresh=\SI{1.0}{\radian}$, which is approximately one third of the total opening angle. 		This additional reward further incentivizes the correct opening motion, leading to faster convergence to good solutions in the training.
		\subsubsection{Body velocity ($\rvel$)}
		To encourage safe behavior, this reward penalizes high body velocities as $\rvel = c_5 \rvelLin + c_6 \rvelAng$, where $\rvelLin = -\norm{\B\pdot}^2$ and $\rvelAng = -\norm{\B\angVel}^2$.
		\subsubsection{Commanded wrench ($\rwrench$)}
		To obtain smooth motions, we penalize high wrench commands with $\rwrench = -c_7 \norm{\wrenchPosContr}^2$.

		\begin{table}[t]
			\renewcommand*{\arraystretch}{1.4}
			\centering
			\begin{tabular}{||c | c c c | c||}
				\hline
				                     & $\bm{x}$              & $\bm{y}$              & $\bm{z}$              & Unit                        \\ [0.5ex]
				\hline\hline
				$\D{\Tilde{\pos}}_d$ & $\Uniform{-0.5}{0.5}$ & $\Uniform{-0.5}{0.5}$ & $\Uniform{-0.5}{0.5}$ & $\SI{}{\centi\meter}$       \\
				\hline
				$\D{\pos}$           & $\Uniform{-0.4}{0}$   & $\Uniform{0}{0.4}$    & $\Uniform{-0.4}{0.4}$ & $\SI{}{\meter}$             \\
				\hline
				$\rotMat_{DB}$       & $\Uniform{-0.2}{0.2}$ & $\Uniform{-0.2}{0.2}$ & $\Uniform{-0.2}{0.2}$ & $\SI{}{\radian}$            \\
				\hline
				$\B\pdot$            & $\Uniform{-0.1}{0.1}$ & $\Uniform{-0.1}{0.1}$ & $\Uniform{-0.1}{0.1}$ & $\SI{}{\meter\per\second}$  \\
				\hline
				$\B\angVel$          & $\Uniform{-0.3}{0.3}$ & $\Uniform{-0.3}{0.3}$ & $\Uniform{-0.3}{0.3}$ & $\SI{}{\radian\per\second}$ \\ [1ex]
				\hline
			\end{tabular}
			\caption{Domain randomization parameters. In the third row ($\rotMat_{DB}$) we interpret the $\bm{x}$, $\bm{y}$, $\bm{z}$ components as RPY angles.\label{tab:domain_rand}}
		\end{table}
		\subsection{Domain Randomization}
		To make the policy's behavior robust, we use domain randomization during training on two main elements:
		\textit{(i)} the door handle position, which is translated from its nominal position on the door with an offset $\D{\Tilde{\pos}}_d$;    \textit{(ii)} the hook initial pose and velocity in the handle frame.
		We randomize these quantities with uniform distributions defined in~\cref{tab:domain_rand}.

		\subsection{Simulation Setup}
		\begin{figure}[t]
			\centering
			\centerline{\includegraphics[width=\columnwidth]{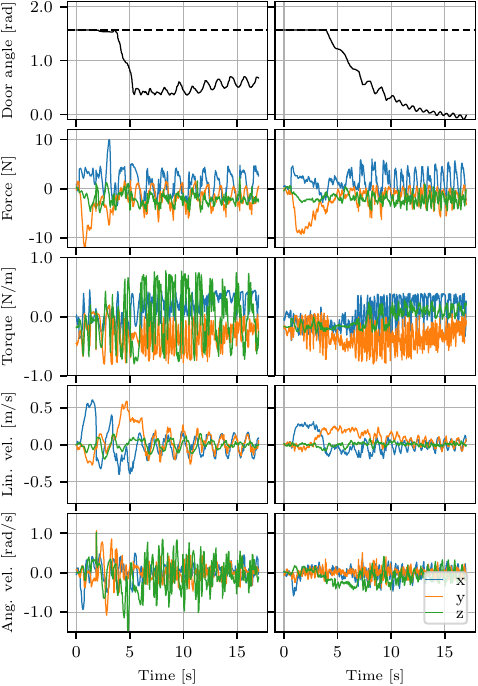}}
			\caption{Results with the zero-shot transferred policy (left column) and retrained with actions saturation (right). From top to bottom: door angle, control forces, control torques, linear and angular velocities in the robot body frame. As can be seen, the initial policy does not succeed in opening the door all the way ($\alpha=0$), while the retrained policy does. The initial policy also has stronger oscillations in the door angle and angular velocities of the body.}
			\label{fig:rl_real}
		\end{figure}
		We train the policy in the Raisim~\cite{raisim} simulator. 		The environment consists of a hinged door with a rigid handle and an \acl{AM} composed of an \ac{OMAV} equipped with a rigid hook. 		We assume the \ac{OMAV} to be a 6-\ac{DoF} rigid floating body on which we can directly apply a command wrench $\wrenchPosContr$. 		We do not include models of the arm's servo motors or spinning propellers in the simulation for two reasons: \textit{(i)} because of the inherent difficulties in identifying and modeling such actuators given backlash, battery voltage dependencies and other nonlinear factors; \textit{(ii)} to keep the simulation complexity as low as possible, allowing the policy to train faster. 				The policy produces pose references that are fed to the robot's pose controller, which is included in the dynamic simulation and generates the command wrench $\wrenchPosContr$.
		\subsection{Training}
		The policy is composed of three fully-connected hidden layers with $256$ neurons each and ReLu activations, as shown in \cref{fig:control_scheme}. The policy is trained by \ac{PPO}~\cite{ppo} inside the Raisim~\cite{raisim} simulation environment. 		We train our policy using 32 CPU cores and an RTX2080 graphics card over \SI{24}{\hour} with $500$ parallel environments, leading to a total of $10$ billion simulated steps.
		\section{Experimental validation}
		Here we present two different evaluations of our approach. First, we transfer the policy to the real robot, discussing the difficulties and the lessons we learned in sim-to-real transfer.
		Second, we compare the \ac{MPPI} and \ac{RL} approaches extensively in simulation, analyzing their robustness to perception mismatches
		and differences in the task initial conditions (closer or farther from the handle).

		\subsection{Real-world experiments}
		Our experimental set-up consists of an \ac{OMAV} and a cabinet in a motion capture system, where the state of both systems (6 \ac{DoF} pose of the robot and $\alpha$ opening angle of the door and its handle's position) are given by motion capture at \SI{200}{\hertz}, as shown in \cref{fig:omav}.
		The motion capture pose estimate is fused with an on-board IMU for pose control of the robot, with our learned policy providing the commanded pose reference at \SI{200}{\hertz}.

		We began our experiments with a zero-shot transferred policy (applying the same policy as was trained in simulation), and discuss its limitations on the real system.
		Second, we attempt to address the sim-to-real gap by training an improved policy.
		The results of both sets of experiments are shown in~\cref{fig:rl_real}.

		\subsubsection{Zero-shot policy transfer}
		The robot starts in free-flight at around \SI{1}{\meter} from the door handle. 		It takes around \SI{4}{\second} to firmly grasp the handle and another \SI{2}{\second} to open the door up to $\doorAngle=\SI{0.4}{\radian}$.
		We notice that the policy can open the door already with a zero-shot transferred policy in only \SI{6}{\second}. 		However, there are two main points to improve: first, the door is not opened all the way to the \SI{0}{\radian} target; second, there are visible oscillations in the motion of the \ac{OMAV}, that become evident once the door is open and reflect into the door's angle. 		The oscillations appear especially in the $\vec{x}$, $\vec{y}$ linear and $\vec{z}$ angular body directions. 		Interestingly, these are the directions of the wrench that mostly require a rotation of the tilting arms of the \ac{OMAV}, which have notably slower dynamics than the propeller motors in the real world.
		These differences were not reflected in our simulation, and our hypothesis is this mismatch causes the oscillations and degraded performance.

		\begin{figure}[t]
			\centering
			\centerline{\includegraphics[width=\columnwidth]{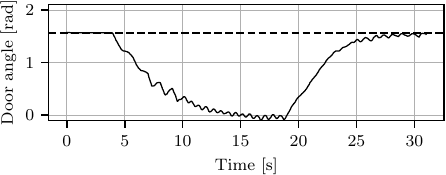}}
			\caption{Door angle opening and closing trajectory. The door is fully closed at $\frac{\pi}{2}$\SI{}{\radian} (indicated by the dashed line) and open at \SI{0}{\radian}. The robot has never seen a door closing scenario during training, but is nevertheless able to smoothly close the door when commanded.}
			\label{fig:open_close}
		\end{figure}
		\subsubsection{Retrained policy}
		To avoid a big excitation of the tilt-arm actuators, we limit the maximum pose displacement that the policy actions can generate by enforcing a maximum norm on the translation offset and a maximum angle offset from the robot's current pose, effectively saturating the policy output.
		The new results are visible in the right column of~\cref{fig:rl_real}. 		The policy is now slower than the previous one due to the new pose command saturations, 		but it reduces the velocity oscillations, smoothly steering the robot and  fully opening the door. 						The policy still induces oscillatory wrench commands but with a smaller amplitude and, interestingly, a higher frequency than before. 		These get partially filtered by the real actuator dynamics, leading to a generally smoother behavior of the system.
		This confirms our hypothesis that the sim-to-real mismatch was largely caused by the lack of actuator dynamics in the simulation.
		\begin{figure}[b]
			\centering
			\centerline{\includegraphics[width=0.9\columnwidth]{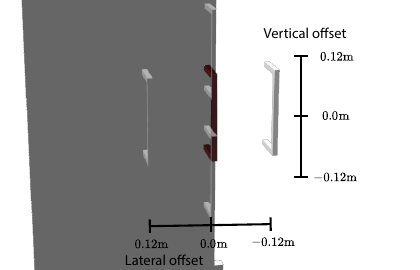}}
			\caption{We intentionally affect the policy's observation with offsets on the door's handle position. The maximum lateral and vertical offsets are shown here. The nominal handle position is in red.}
			\label{fig:handle_translation}
		\end{figure}
		\begin{figure*}[t]
			\centering
			\centerline{\includegraphics[width=\textwidth]{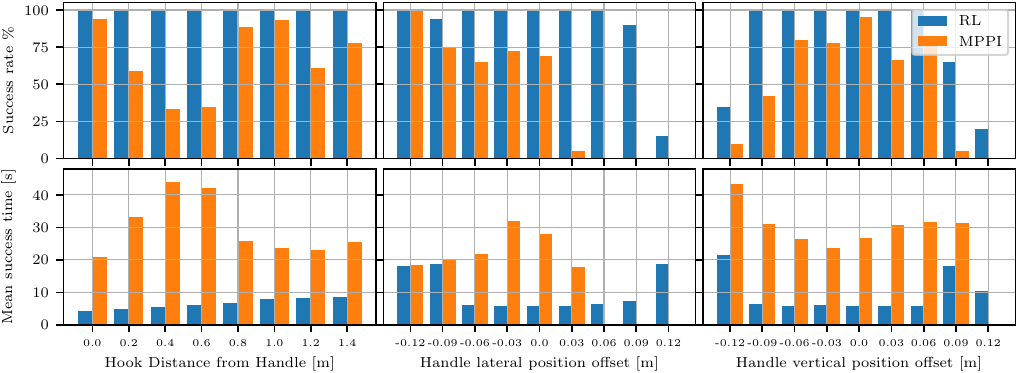}}
			\caption{Door opening success rate comparison between \ac{RL} policy and \ac{MPPI} controller. The top row shows the success rates in three different experiments, while the lower row contains the respective mean amounts of time to complete the task. From left to right, the three experiments are respectively: initializing the robot's position further and further from the door, degrading the handle's position observation with lateral offsets and with vertical offsets. Missing bars mean that there was not one success in that scenario. As a main result, the policy is consistently faster and more successful than \ac{MPPI} at solving the task.}
			\label{fig:mppi_comparison}
		\end{figure*}
		A video of the real-world experiments can be found at \href{https://youtu.be/5-2aTli3au0}{youtu.be/5-2aTli3au0}.

		The \ac{RL} policy has a number of desirable emergent properties: first, it's able to recover from failures and make multiple attempts at opening the door.
		This robustness is further analyzed in \cref{sec:robustness}.
		Second, the same policy is also able to close the door even if it never experienced this scenario in simulation. 		By changing the door angle observation from $\doorAngle$ to $\doorAngle - \doorAngle_\text{target}$ during deployment and setting $\doorAngle_\text{target}=\frac{\pi}{2}\SI{}{\radian}$, the robot successfully closes the door, as shown in~\cref{fig:open_close}.

		This shows the ability of our policy to generalize to other variations on the same task, and that it doesn't simply memorize a desirable sequence of actions during training.

		\subsection{Robustness and comparison to related work}
		\label{sec:robustness}
		To further evaluate the effect of domain randomization on the robustness of the resulting policy, and to compare to a non-\ac{RL}-based approach, we conduct a series of simulation experiments.

		Here we compare our learned policy with the recent \ac{MPPI}-based door opening method from~\cite{brunner2022planning} and  evaluate the robustness of both approaches to errors in the model of the environment and changes in initial conditions.
		The robot's only knowledge of the state of the world consists of the position of the door handle $\B\pos_{hd}$ and the door plane $\frameD$, both relative to the \ac{OMAV}, and the door opening angle $\alpha$.
		We want to evaluate the impact of errors in the estimation of these variables, which are likely on the real system due to calibration errors or sensor noise.

		The three experiments performed to evaluate these are:
		\begin{itemize}
			\item Changing the initial conditions by increasing the hook distance from the handle.
			\item Introducing an offset error on the observation of the handle's position along the $\bm{z}_D$ axis of $\frameD$ (\cref{fig:handle_translation}).
			\item Introducing an offset error on the observation of the handle's position along $\bm{y}_D$ axis of $\frameD$.
		\end{itemize}

		Similar to the real-world set-up, the goal is to open the door to an angle of \SI{90}{\degree}. 		The task is considered successful when the door stops in the range between $\pm 10\%$ from the target angle, or failed if \SI{60}{\second} passed without reaching the success condition.
		In all three test scenarios, we repeat the door-opening task 20 times for each increment and each of the two methods, for a total of $1040$ different task executions over \SI{24}{\hour} of data collection. 		We use Gazebo as the simulation environment, which is different from the Raisim training environment of the policy and contains more realistic flight dynamics and sensor noise. 				The results of the experiments are in \cref{fig:mppi_comparison}.
		\\
		\subsubsection{Hook distance to the handle}
		Here we progressively increase the initial distance of the robot's hook from the door handle, starting from \SI{0.0}{\meter} where the hook is almost inside the handle, up to a distance of \SI{1.4}{\meter}. 		While the distance is fixed at a specific radius for every step, the actual position of the robot is randomly sampled.
		The left column of \cref{fig:mppi_comparison} shows how the distance does not affect the task success rate for the policy, achieving an astonishing $100\%$ at every distance. 		This demonstrates the benefit of the domain randomization in initial positions that the policy was trained with.
		Even if it experienced training distances only up to \SI{0.4}{\meter} and only on the right side of the handle (along the positive $\bm{y}_D$ axis of $\frameD$), the policy is still successful at greater distances. 		On the contrary, the \ac{MPPI} approach shows very good performance when very close to the handle, which slightly worsens when increasing the distance, getting better again after a distance of \SI{0.8}{\meter}. 		This is likely due to the momentum that the robot obtains during the approach phase, helping the system to better hook into the door's handle.
		The reinforcement learned policy is also significantly faster, successfully opening the door in less than \SI{10}{\second} in all cases, while \ac{MPPI} requires between \SI{20}{\second} and \SI{45}{\second}.
		The reason is that \ac{MPPI} would often spend most of the time trying to fit the hook into the door's handle, while the policy would generate more effective hooking trajectories.
		\\
		\subsubsection{Offset on the handle lateral $y_D$ position observation}
		Here we progressively add an offset to the door handle $\bm{y}_D$  observation from \SI{-0.12}{\meter} to \SI{0.12}{\meter}.
		Since the initial position of the robot hook is always the same on the right side of the handle, a negative offset moves the real handle position closer to the hook, while a positive offset moves it further away than the controller observes. 		In case the handle is closer, both controllers perform well, with performance degrading as the handle gets further and further away. 		The policy has a better behavior than the \ac{MPPI}, only failing after a \SI{10}{\centi\meter} displacement.
		Comparing task completion times, the policy is again the fastest, with time only increasing when the handle has large observation errors.
		\\
		\subsubsection{Offset on the handle vertical $z_D$ position observation}
		Here we add offsets to the $\bm{z}_D$ position observation. A negative offset means the handle is shifted down, while a positive offset means the handle is at a higher position. Both controllers perform very well while the handle is around the nominal position, decreasing the performance as it moves up or down. Again the policy exhibits a more robust behavior.

In summary, the learned policy is more robust to errors in its observations and changing initial conditions than \ac{MPPI}. We believe this robustness to come from the domain randomization that the policy experienced during training.
Interestingly, even if the door handle position offset was randomized only \SI{0.5}{\centi\meter}, the policy shows to be robust for errors up to \SI{9}{\centi\meter}.
Our policy is also much faster to evaluate than \ac{MPPI}, running at \SI{200}{\hertz} against \ac{MPPI}'s \SI{10}{\hertz}.

\section{Conclusions}
This work presented an \ac{RL} approach to the problem of articulated object manipulation with fully actuated aerial robots. We were able to solve a complex door opening task without the need for an accurate analytical model or computationally heavy online optimizations.
Our policy was able to transfer well from simulation to real world with only minimal modifications, and succeeded not only in opening the door, but also in closing it, which is a task it never witnessed during training.
This suggests that these approaches can be used to generalize to wider varieties of tasks.
We compared our approach to a recent \ac{MPPI} solution for the door-opening task, showing an increase in robustness and performance with respect to the state of the art. In particular, the policy succeeded at levels of degraded observations in which the model-based solution would completely fail. Future works will address the problem of simulation to reality transfer of more dynamic motions, giving the policy more freedom in exploring door-opening strategies.
We believe that \acl{RL} is capable of finding robust, computationally-efficient, and generalizable solutions to complex aerial manipulation tasks.

\section*{ACKNOWLEDGMENT}
We thank Helen Oleynikova for her help and thorough review of the final version of this work.

\bibliographystyle{IEEEtran}
\bibliography{references}

\end{document}